\documentclass[utf8]{FrontiersinHarvard} 

\usepackage{url,hyperref,lineno,microtype,subcaption}
\usepackage[onehalfspacing]{setspace}

\usepackage{multirow}
\usepackage{booktabs}

\def\keyFont{\fontsize{8}{11}\helveticabold }
\def\firstAuthorLast{Wu {et~al.}} 
\def\Authors{
Meihong Wu\,$^{1}$,
Xiaoyan Cao\,$^{1}$ 
and Shihui Guo\,$^{1,*}$
}
\def\Address{
$^{1}$School of Informatics, Xiamen University, Xiamen, Fujian, China
}
\def\corrAuthor{School of Informatics, Xiamen University, Binhai Street, Siming District, Xiamen City, Fujian Province, China, 361005}

\def\corrEmail{guoshihui@xmu.edu.cn}

\begin{document}
\onecolumn
\firstpage{1}

\title[Swarm behavior tracking based on a deep vision algorithm]{Swarm behavior tracking based on a deep vision algorithm} 

\author[\firstAuthorLast ]{\Authors} 
\address{} 
\correspondance{} 

\extraAuth{}

\maketitle

\begin{abstract}


The intelligent swarm behavior of social insects (such as ants) springs up in different environments, promising to provide insights for the study of embodied intelligence.
Researching swarm behavior requires that researchers could accurately track each individual over time.
Obviously, manually labeling individual insects in a video is labor-intensive. Automatic tracking methods, however, also poses serious challenges: (1) individuals are small and similar in appearance; (2) frequent interactions with each other cause severe and long-term occlusion.
With the advances of artificial intelligence and computing vision technologies, we are hopeful to provide a tool to automate monitor multiple insects to address the above challenges.
In this paper, we propose a detection and tracking framework for multi-ant tracking in the videos by: (1) adopting a two-stage object detection framework using ResNet-50 as backbone and coding the position of regions of interest to locate ants accurately; (2) using the ResNet model to develop the appearance descriptors of ants; (3) constructing long-term appearance sequences and combining them with motion information to achieve online tracking. 
To validate our method, we construct an ant database including 10 videos of ants from different indoor and outdoor scenes.
We achieve a state-of-the-art performance of 95.7\% mMOTA and 81.1\% mMOTP in indoor videos, 81.8\% mMOTA and 81.9\% mMOTP in outdoor videos.
Additionally, Our method runs 6-10 times faster than existing methods for insect tracking.
Experimental results demonstrate that our method provides a powerful tool for accelerating the unraveling of the mechanisms underlying the swarm behavior of social insects.
  
\tiny
 \keyFont{ \section{Keywords:} social insects, swarm behavior, automatic tracking method, object detection, online tracking} 
\end{abstract}

\section{Introduction}

Swarm behavior is one of the most important features of social insects, which has important significance for the study of embodied intelligence ~\cite{tiacharoen2012design}. Specifically, social insects often tend to cluster into a colony~\cite{vandermeer2008clusters}, which forms a complex dynamical system together with the surrounding environment~\cite{balch2001automatically}.
So far, we do not know enough about the mechanisms behind swarm behaviors of social insects.

The mainly reason is that the key requirement of this research is the ability to track the motions and interactions of each individual robustly and accurately. 
However, until the late $20^{th}$ century, biologists still manually marked the motion trajectories on the video to guarantee the quality. They have to track each individual at one time, which might mean watching the entire video 50 times or more in a crowded scene~\cite{poff2012efficient}. 
It becomes an inhibiting factor in obtaining the complete and accurate dataset required to analyze the evolution of complex dynamical system.
Therefore, in the past two decades, attempts have been made to automate the tracking process for social insects utilizing computer vision (CV) techniques~\cite{khan2005mcmc, khan2006mcmc, oh2006parameterized, veeraraghavan2008shape, fletcher2011multiple}.

Traditional CV techniques free researchers from manual work through approaches such as foreground segmentation algorithm~\cite{li2008estimating}, temporal difference method~\cite{khan2005mcmc} and hungarian algorithm~\cite{li2009learning}.
Such approaches, however, have failed to address the noise in the image~\cite{zhao2015improved}, resulting in the limitation that a laboratory environment with a clean background is needed.
Nevertheless, many scientifically valuable results are obtained in nature rather than laboratory environment~\cite{schmelzer2009special, kastberger2013social, tan2016honey, dong2018olfactory}.

In recent years, with the popularity of computer vision, many advanced object detection and tracking methods have emerged.

\subsection{Object detection}
Existing methods in object detection are categorized as one-stage or two-stage, according to whether there is a separate stage of region proposal.
One-stage frameworks (e.g., YOLO~\cite{redmon2016you}) are fast, but their accuracy is typically slightly inferior compared with that of two-stage detection.
The popularity of two-stage detection frameworks is enhanced by R-CNN~\cite{girshick2014rich}, which proposes candidate regions via a selective search (SS) algorithm~\cite{uijlings2013selective}, thereby the detector focuses on these RoIs. 
However, using the SS algorithm~\cite{uijlings2013selective} to generate region proposals is the main reason causing slow inference.
Fast R-CNN~\cite{girshick2015fast} reduces the computational complexity of region proposals by downsampling the original image, while Faster R-CNN~\cite{ren2015faster} proposes an RPN, which further improves the speed of training and inference. 

Given the success of deep learning in general tasks of object detection, researchers also applied to detect specific groups of animals, such as a single mouse~\cite{geuther2019robust}, fruit flies~\cite{murali2019classification}.
These methods are either limited to track a single object, or a fixed number of objects.
General tools~\cite{romero2019idtracker,sridhar2019tracktor} also offer the functionality to detect and track unmarked animals in the image.
However, most of existing methods focus on the condition of ideal lab set-up and none of existing works reported the detection of ants in outdoor environments which contain diverse backgrounds and arbitrary terrains.


\subsection{Multi-object tracking (MOT)}
In the last two decades, vision-based detection and tracking models have been widely used to study social insects~\cite{khan2005mcmc,veeraraghavan2008shape}. 
Appearance (particularly color) and motion information are the main metrics used in this category of method. 
Due to high similarity of ants’ appearance, researchers either use the technique of pigmenting to create more distinct appearance features~\cite{fletcher2011multiple}, or limit the observation to a laboratory setup~\cite{branson2009high, perez2014idTracker}.
State-of-the-art methods, such as Ctrax~\cite{branson2009high} and idTracker~\cite{perez2014idTracker}, for insect tracking are tested in a laboratory setup and use background subtraction for foreground segmentation. 
Notably, the operations of background modeling and foreground extraction are time-consuming.

The tracking-by-detection (TBD) paradigm is to match trajectories and detections in two consecutive frames, a process that requires metrics. 
The global nearest neighbor model measures motion state to achieve Drosophila tracking~\cite{chiron2013detecting}. 
The global nearest neighbor model assumes that the motion state obeys the linear observation model, which commonly uses a constant velocity model - the Kalman filter (KF). 
However, changes in ants’ speed and direction are difficult to predict, thus appearance information is integrated as a metric.

The DAT method is a mainstream method for ant colony tracking~\cite{fasciano2014ant}. 
It allows a combination of multiple metrics, and uses Hungarian algorithm~\cite{li2009learning} to assign detections for trajectories.
The PF method is suitable for solving nonlinear problems~\cite{fasciano2013tracking}, but the growth in the number of particles leads to an exponential increase in the computational cost, preventing the effective multi-object tracking. Using Markov Chain Monte Carlo sampling can reduce computational complexity~\cite{khan2006mcmc}. 
A GPU-accelerated semi-supervised framework can further improve tracking accuracy and performance~\cite{poff2012efficient}.

When applying the methods above for tracking ant colonies, they are greatly disturbed by background noise and difficult to overcome the serious occlusion problem in dense scenes. 
Long short-term memory~\cite{kim2018multi} and spatial-temporal attention mechanisms~\cite{zhu2018online} have been developed to tackle the problem of long-term occlusion.
A bilinear Long short-term memory structure that couples a linear predictor with input detection features, thereby modeling long-term appearance features~\cite{kim2018multi}. 
The spatial-temporal attention mechanism is also suitable for the MOT task. 
The spatial attention module makes the network focus on the pattern of matching. Meanwhile the temporal attention module assigns different levels of attention to the sample sequence of the trajectory~\cite{zhu2018online}. 
The TBD paradigm-based framework is dependent on detection results. Therefore, severe occlusion is likely to cause tracking failures. 
To prevent this situation, a detector with automatic bounding box repairing and adjustment is introduced by a cyclic structure classifier~\cite{dong2016occlusion}.

\begin{figure}[!htbp]
	\centering
	\includegraphics[width=\linewidth]{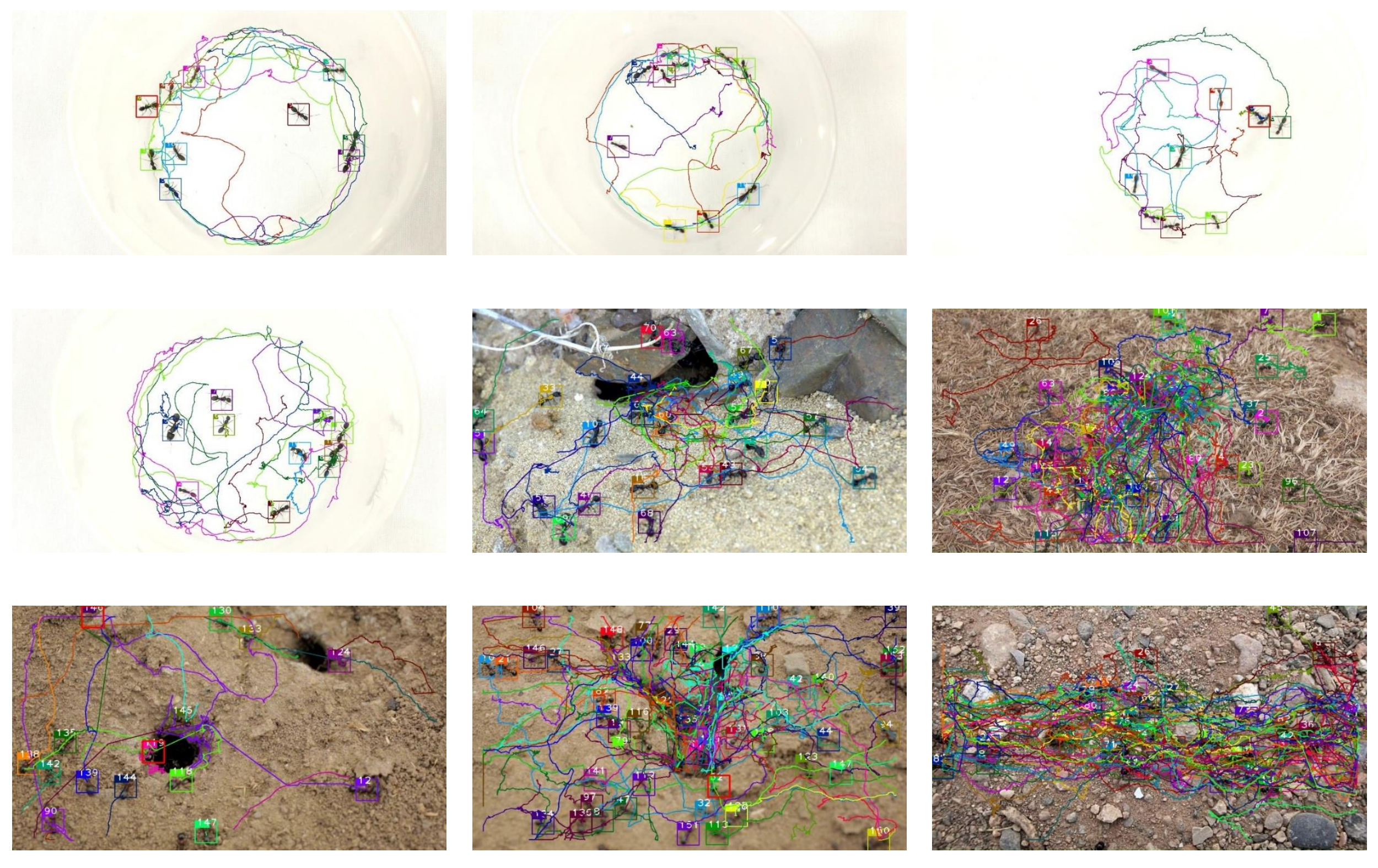}
	\caption{\label{fig:tracking_results} Tracking results by our method in both indoor and outdoor environments.}
\end{figure}

In this paper, we use a deep learning method to build a detection and tracking framework.
Our method is based on the TBD paradigm and accomplishes the goal of online multi-ant tracking.
To the best of authors' knowledge, this is the first work to achieve robust detection and tracking of ant colony in both indoor and outdoor environments (Figure~\ref{fig:tracking_results}).
Our method is robust in tackling the challenge of visual similarity among colony individuals, handling diverse terrain backgrounds and achieving long-period of tracking.
Our main contributions are as follows:
\begin{itemize}
	\item We adopt a two-stage object detection framework, using ResNet-50 as the backbone and position sensitive score maps to encode regions of interest (RoIs).
	During the tracking stage, we use a ResNet network to obtain the appearance descriptors of ants and then combine them with motion information to achieve online association.
	
	\item Our method proves to be robust in both indoor and outdoor scenes. Furthermore, only a small amount of training data are required to achieve the goal in our pipeline, which are 50 images chosen for each scene in the detection framework and 50 labels randomly chosen for the tracking framework respectively.
	
	\item We construct an ant database with labeled image sequences, including five indoor videos (laboratory setup) and five outdoor videos, with 4983 frames and 115,433 labels in total. 
	The database is made publicly available, which is hoped that it will contribute to a deeper exploration on swarm behavior of ant colony.
\end{itemize}

\section{Materials and methods}
\subsection{Overview}
\label{sec:overview}
\begin{figure}[!htbp]
	\centering
	\includegraphics[width=0.65\linewidth, height=0.7\linewidth]{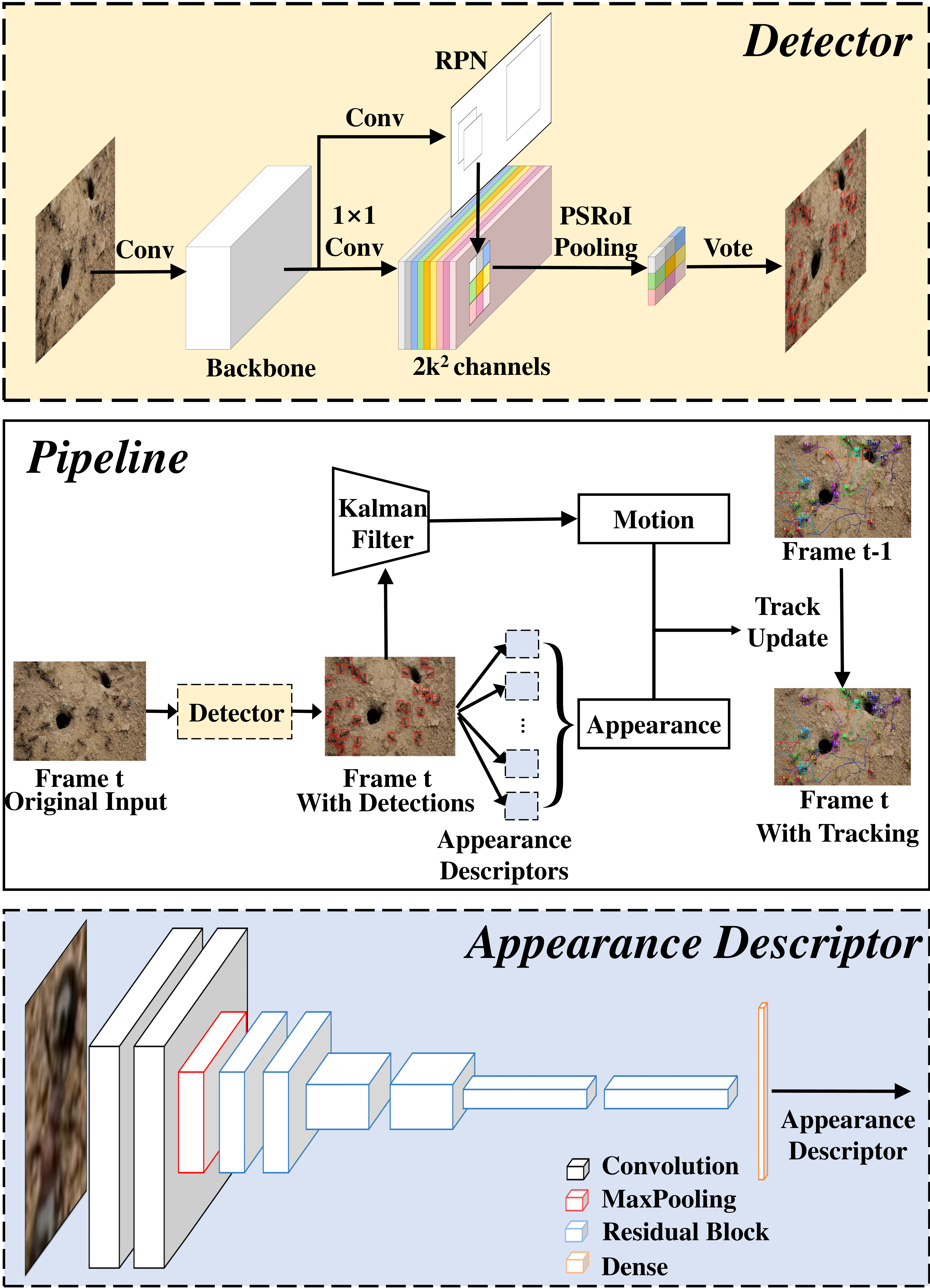}
	\caption{\label{fig:pipeline} Architecture for detection and tracking.}
\end{figure}

Following the TBD paradigm, we propose a uniform framework for detection and tracking to efficiently and accurately track the ant colony in both indoor and outdoor scenes (Figure~\ref{fig:pipeline}). 
In the detection phase, we adopt a two-stage object detection framework, using ResNet-50 as the backbone, and encoding RoIs proposed by regional proposal network via position-sensitive score maps. 
Then we implement classification and regression through downsampling and voting mechanisms.
(see details in Section~\nameref{sec:two-stage}). 
In the tracking stage, we first use ResNet to train the appearance descriptors of ants and measure the appearance similarity between two objects. 
Next, the tracking is accomplished by combining appearance and motion information for online association metric. (Section~\nameref{sec:MOT_framework}).

\subsection{Two-stage object detection}
\label{sec:two-stage}
\subsubsection{Regional proposal network}
Regional proposal network (RPN) is proposed in Faster R-CNN \cite{ren2015faster} to generate RoIs. 
Compared to SS \cite{uijlings2013selective}, RPN is based on the CNN network structure and can connect the backbone with shared weight, significantly improving detection speed. 
We use ResNet-50 as the backbone and replace the fully connected layer with a 1$\times$1 convolution to reduce the dimensions of feature maps.
Considering that ResNet-50 conducts downsampling 32 times, we get 256-d feature maps via a 3$\times$3 Atrous convolution to maintain translation variability.
For each sliding position, we predict k region proposal boxes of different sizes and ratios; these boxes are called anchors.
After the 256-d vector, we connect classification and regression branches through two parallel 1$\times$1 convolution layers. 
The classification branch uses softmax to determine whether there is an object in anchor so that this branch has 2$\times$k outputs. 
The regression branch will perform a regression on the 4D position parameters of anchors (i.e., center coordinates, width and height) so that there are 4$\times$k outputs. 
RPN will propose k$\times$w$\times$h anchors with a w$\times$h feature map, called RoIs. 
We use the Non-maximum suppression algorithm \cite{neubeck2006efficient} to filter duplicate anchors and set the IOU threshold to 0.7.

\subsubsection{Position sensitive region of interest}
On the basis of RoIs, the two-stage detection framework classifies and fine-tunes the location of bounding boxes.
In Faster R-CNN, RoIs are scaled to the last feature maps and focusing on these areas through ROIPooling.
Next, each RoI is classified and regressed through two fully connected layers, causing high computational complexity.

In order to reduce the number of parameters, we use RPN-FCN \cite{dai2016r} to generate position-sensitive score maps via a convolutional layer, which is connected to the backbone. 
Both classification and regression tasks have independent position-sensitive score maps, forming three parallel branches with RPN.

For the classification task, since we only need to classify ants and background, we use k$\times$k$\times$2 convolution kernels to generate score maps.
k$\times$k indicates that each RoI is divided into k$\times$k regions to encode position information. 
Each region is encoded by a specific feature map with two dimensions.
Similarly, we use k$\times$k$\times$4 convolution kernels for fine-tuning the position of RoIs in the regression task.
 
To focus on RoIs, we perform average pooling on each region to get feature maps, called position sensitive region of interest (PSRoI) pooling, as the following formula shows:

\begin{equation}
r_{c}(i, j | \Theta)=\sum_{(x, y) \in \text {region}(i, j)} z_{i, j, c}\left(x+x_{0}, y+y_{0} | \Theta\right) / n.
\end{equation}
$r_{c}(i, j | \Theta)$ is the result of downsampling in $(i,j)^{th}$ for $c^{th}$ category, and $z_{i, j, c}$ is one score map in the k$\times$k$\times$2 position-sensitive score maps. $(x0, y0)$ represents the left-top corner of RoI. $\Theta$ is the set of parameters of the network, and $n$ is the number of pixels in the region.

For the feature maps, we vote on k$\times$k regions, getting the overall score of RoI on the classification or regression task, as the following formula shows:

\begin{equation}
r_{c}(\Theta)=\sum_{i, j} r_{c}(i, j | \Theta).
\end{equation}
In the formula, $r_{c}(\Theta)$ represents the overall scores of all regions.

Next, we use softmax to implement binary classification, as the following formula shows:

\begin{equation}
s_{c}(\Theta)=e^{r_{c}(\Theta)} / \sum_{c=0}^{C} e^{r_{c}(\Theta)}.
\end{equation}
Here, $s_{c}(\Theta)$ is the probability of $c^{th}$ category.
Finally, we use the Non-maximum suppression algorithm to filter the bounding box.

Since object detection includes classification and regression, we require a multitask loss function. 
In this paper, we weight the loss functions of the two tasks. Because softmax is used for the binary classification task, it is natural to adopt cross-entropy loss for the classification task. For the regression task, we calculate the matching degree between the four position parameters and ground truth:
\begin{equation}
L\left(s, t_{x, y, w, h}\right)=L_{c l s}\left(s_{c^{*}}\right)+\lambda\left[c^{*}=1\right] L_{r e g}\left(t, t^{*}\right).
\end{equation}
where $c^{*}$ is the ground truth category label of RoI, and $c^{*} = 1$ represents ants.
$L_{c l s}\left(s_{c^{*}}\right)$ represents cross-entropy loss:
\begin{equation}
L_{c l s}\left(s_{c}\right)=-\log \left(s_{c}\right).
\end{equation}
$L_{r e g}\left(t, t^{*}\right)$ represents the loss of the regression task, including 4 dimensions:
\begin{equation}
L_{r e g}\left(t, t^{*}\right)=\Sigma_{i=1}^{4} g\left(t_{i}^{*}-t_{i}\right).
\end{equation}
In the formula, $t^{*}$ is the predicted position, and $t$ is ground truth after translation and scaling.

\subsection{MOT framework}
\label{sec:MOT_framework}
\subsubsection{Offline ResNet network architecture}
We adopt a 15-layer ResNet network architecture to extract the appearance descriptors of objects, as Figure~\ref{fig:pipeline} shows. 
After downsampling eight times, the network will eventually obtain a 128-dimensional feature vector through a fully connected layer. The specific parameters are consistent with \cite{CAO2020107233}.

\subsubsection{Cosine similarity metric classifier}
We modify the parameters of softmax to get a cosine similarity measurement classifier, which can measure the similarity of the same category or different categories. First, the output of a fully connected layer is normalized by batch normalization, ensuring that it is expressed as a unit length $\left\|f_{\Theta}(x)\right\|_{2}=1$, $\forall x \in R^{D}$. Second, we normalize the weights, that is, $\varpi_{k}=\omega /\left\|\omega_{k}\right\|_{2}$, $\quad\forall k=1, \ldots C$. 
Cosine similarity metric classifier is constructed as follows:

\begin{equation}
p\left(y_{i}=k | r_{i}\right)=\frac{\exp \left(\kappa \cdot \varpi_{k}^{T} r_{i}\right)}{\sum_{n=1}^{C} \exp \left(\kappa \cdot \varpi_{n}^{T}\right)}.
\end{equation}
Here, $\kappa$ is the free scaling parameter.

Because the cosine similarity classifier follows the structure of softmax, we use the cross-entropy loss for training:

\begin{equation}
L(D)=-\sum_{i=1}^{N} \sum_{k=1}^{C} \mathrm{gt}_{y_{i}-k} \cdot \log p\left(y_{i}=k | r_{i}\right).
\end{equation}
Here, $L(D)$ represents the sum of the cross-entropy loss of $N$ images, $p\left(y_{i}=k | r_{i}\right)$ is the prediction result of $i^{th}$ image in $k^{th}$ label, and $\mathrm{gt}_{y_{i}-k}$ is ground truth.

\subsubsection{Motion matching}
We use the KF model to predict the position of trajectories in the current frame. 
Then, we calculate the square of the Mahalanobis distance between the predicted position and the detected bounding box position by measuring the degree of motion matching \cite{wojke2017simple} as follows:

\begin{equation}
d^{(1)}(i, j)=\left(d_{j}-y_{i}\right)^{T} S_{i}^{-1}\left(d_{j}-y_{i}\right).
\end{equation}
Here, $d_{j}$ is the position of the $j^{th}$ detection box, $y_{i}$ is the position of the $i^{th}$ trajectory predicted by the KF, and $S_{i}$ is the covariance matrix between the $i^{th}$ trajectory and the detected bounding box.

We use a 0-1 variable to indicate whether trajectory and detection meet the association conditions. If the Mahalanobis distance meets $t^{\left(1\right)}$, $\left(i,j\right)$ will be added to the association set. The formula can be expressed as:
\begin{equation}
b_{i j}^{(1)}=\left\{\begin{array}{ll}
{1,} & {d^{(1)}(i, j)<t^{(1)}} \\
{0,} & {\text{ otherwise }}
\end{array}\right..
\end{equation}
Here, $b_{i j}^{(1)}$ is the motion association signal.

\subsubsection{Appearance matching}
\label{sec:appearance_matching}
We use the appearance descriptors to measure the appearance similarity between ants. Furthermore, we create a gallery for each trajectory, and each gallery stores the latest 100 appearance descriptors. 
Then, we calculate the cosine distance of appearance descriptors between gallery and candidate bounding boxes. 
The smallest distance is used as an appearance matching degree as follows:

\begin{equation}
d^{(2)}(i, j)=\min \left\{1-r_{j}^{T} r_{k}^{(i)} | r_{k}^{(i)} \in \mathrm{K}_{i}\right\}.
\end{equation}
where $r_{j}$ is the appearance descriptor of the $j^{th}$ detection box, $r_{k}^{(i)}$ is the $k^{th}$ appearance descriptor of the $i^{th}$ trajectory, $d^{(2)}(i, j)$ represents the appearance matching degree between the $i^{th}$ trajectory and the $j^{th}$ bounding box.

Similarly, we introduce a 0-1 variable as an association signal. If the appearance matching degree from a pair of trajectory and detection boxes meets the threshold, we add it to the association set:
\begin{equation}
b_{i j}^{(2)}=\left\{\begin{array}{ll}
{1,} & {d^{(2)}(i, j)<t^{(2)}} \\
{0,} & {\text { otherwise }}
\end{array}\right..
\end{equation}
where $b_{i j}^{(2)}$ represents the appearance association signal. In this paper, $t^{(2)}$ is set to 0.2.

\subsubsection{Comprehensive matching}
To combine motion and appearance information, we set a comprehensive association signal $b_{i j}$. Only when both motion and appearance matching degree meet the threshold, the $\left(i, j\right)$ pair will be considered for matching. The formula expression is denoted as follows:

\begin{equation}
b_{i j}=\prod_{m=1}^{2} b_{i, j}^{(m)}.
\end{equation}

However, the KF is scarcely possible to track accurately for long periods, because of the motion of ants is complicated.
Therefore, we use the appearance matching degree (Section~\nameref{sec:appearance_matching})
as the association cost.

\subsubsection{Track update}
First, we use matching cascade to match in priority for the most recently associated trajectories, avoiding the trajectory drift caused by long-term occlusion \cite{wojke2017simple}. 
During the matching, we use the Hungarian algorithm to find the minimum cost matches in the association cost matrix. 
For unmatched trajectories and detection boxes, we calculate the IOU. 
If they meet the threshold, they are associated.
After that, trajectories need to be updated. They have three states: unconfirmed, confirmed, and deleted. We assign a new trajectory for each unmatched detection. 
Furthermore, if duration of trajectory is less than three, it will be set to an unconfirmed state.
The unconfirmed trajectories need to be successfully associated for three consecutive frames before being converted into confirmed state; otherwise, they will be deleted.
For the unmatched confirmed trajectories, if they are successfully matched in the previous frame, the KF will to estimate and update their motion state in the current frame; otherwise, we will suspend tracking.
Moreover, if the number of consecutively lost frames of confirmed trajectories exceeds the threshold (Amax=30), they will be deleted.


\section{Results}
\label{sec:experiment}

\subsection{Ant colony database}
\label{sec:database}

We establish an video database of ant colony, which contains a total of 10 videos. 
Five videos are from an existing published work \cite{CAO2020107233} and captured in the indoor (laboratory) environment.
The remaining five outdoor videos are captured in different backgrounds and are obtained from the online website \emph{DepositPhotos} (http://www.depositphotos.com).
Table~\ref{tab:Statistics} shows detailed video information, where $\mathcal{I}$ represents an indoor video,  $\mathcal{O}$ represents an outdoor video.
The resolutions of indoor and outdoor videos are 1920$\times$1080 and 1280$\times$720, respectively.

\begin{table}[tbhp]
\centering
\begin{tabular}{|c|c|c|c|c|c|}
\hline
\textbf{Sequence} & \textbf{FPS}        & \textbf{Resolution}        & \textbf{Length} & \textbf{Ants} & \textbf{Annotations} \\ \hline
$\mathcal{I}_1$           & \multirow{5}{*}{25} & \multirow{5}{*}{1920$\times$1080} & 351 (00:14)     & 10              & 3510           \\ \cline{1-1} \cline{4-6} 
$\mathcal{I}_2$           &                     &                            & 351 (00:14)     & 10              & 3510           \\ \cline{1-1} \cline{4-6} 
$\mathcal{I}_3$           &                     &                            & 351 (00:14)     & 10              & 3510           \\ \cline{1-1} \cline{4-6} 
$\mathcal{I}_4$           &                     &                            & 351 (00:14)     & 10              & 3510           \\ \cline{1-1} \cline{4-6} 
$\mathcal{I}_5$           &                     &                            & 1001 (00:40)    & 10              & 3510           \\ \hline
$\mathcal{O}_1$           & \multirow{5}{*}{30} & \multirow{5}{*}{1280$\times$720}  & 600 (00:20)     & 73              & 11178          \\ \cline{1-1} \cline{4-6} 
$\mathcal{O}_2$           &                     &                            & 677 (00:23)     & 162             & 25158          \\ \cline{1-1} \cline{4-6} 
$\mathcal{O}_3$           &                     &                            & 577 (00:19)     & 133             & 10280          \\ \cline{1-1} \cline{4-6} 
$\mathcal{O}_4$           &                     &                            & 526 (00:18)     & 193             & 27902          \\ \cline{1-1} \cline{4-6} 
$\mathcal{O}_5$           &                     &                            & 569 (00:19)     & 101             & 22044          \\ \hline
\end{tabular}
\caption{Statistics of ant videos with annotations in indoor and outdoor scenes.}
\label{tab:Statistics}
\end{table}

The videos in our database have a total of 4983 frames. 
There are 10 ants per frame in the indoor videos. 
The number of ants in each frame is 18-53 in the outdoor videos.
The number of objects in this scenario is significant, considering the fact that the popular COCO benchmark dataset contains only on average 7.7 instances per image. 
Some video characteristics present challenges for detection and tracking algorithms, for example over-exposure for indoor videos and diverse background for outdoor ones.

There are caves or rugged terrains in outdoor scenes, and ants may enter or leave the scene.
Different from multi-human tracking, ants are visually similar and this causes significant challenges for tracking.
We manually mark the video frame by frame. 
To facilitate training and reduce labeling cost, the aspect ratio of each bounding box is 1:1. 
Considering the posture and scale of ants, we set the size of the bounding box to 96$\times$96 for indoor videos and 64$\times$64 for outdoor videos. The database and code will be made publicly available.

\subsection{Evaluation index}
\label{sec:evaluation}

In this paper, the evaluation indicators of detection and tracking performance are as follows:

\begin{itemize}

	\item Mean Average Precision (MAP): the weighted sum of the average precision of all videos. The weight value is the proportion of frames.
	
	\item False Positive (FP): the total number of false alarms.
	
	\item False Negative (FN): the total number of objects that do not match successfully.
	
	\item Identity Switch (IDS): the total number of identity switches during the tracking process.
	
	\item Fragments (FM): the total number of incidents where the tracking result interrupts the real trajectory.
	
	\item mean Multi-object Tracking Accuracy (mMOTA): the weighted sum of the average tracking accuracy of all videos. The equation to compute mMOTA is: mMOTA = 1 - (FP + FN + IDS)/NUM\_LABELED\_SAMPLES, where NUM\_LABELED\_SAMPLES is the total number of labeled samples.
	
	\item mean Multi-object Tracking Precision (mMOTP): the weighted sum of the average tracking precision of all videos. Tracking precision measures the intersection over union (IOU) between labeled and predicted bounding boxes.
	
	\item Frame Rate (FR): the number of frames being tracked per second.
	
\end{itemize}


\subsection{Results of multi-ant detection}
In our ant database, we set up five groups of training sets (Table~\ref{tab:det_results}) and compare their performance with that of the remaining datasets. 
The naming conventions are:

\begin{itemize}
	\item 
	$\mathcal{I}_5$+$\mathcal{O}_4$ represents a union of the $5^{th}$ indoor video and the $4^{th}$ outdoor video.
	\item
	$\mathcal{I}_{1-4}$ represents a union of indoor videos with their IDs of [1,2,3,4].
	\item 
	$\mathcal{I}_5\left(50\right)$ represents the last 50 frames selected from the $5^{th}$ indoor video. This partition strategy ensures the frame continuity for the subsequent tracking task.
	\item
    $\mathcal{O}_{1-5}\left(-50\right)$ represent the union of 5 subsets, the last 50 frames de-selected from the outdoor videos with their IDs of [1,2,3,4,5].
\end{itemize}

In all scenarios, the detection accuracy of indoor videos is higher than that of outdoor videos, and MAP reaches over 90\%. 
We also noticed that the test result for outdoor videos was only 49.7\% on $\mathcal{I}_5$+$\mathcal{O}_4$. 
This is because we used only $\mathcal{O}_4$ as the outdoor training set, which is insufficient to cover the wide range of diversity in terms of environmental backgrounds and ant appearances.

\begin{table}[!h]
	\centering
	\begin{tabular}{@{}llccc@{}}
		\toprule
		\multicolumn{1}{c}{Training Data}   & \multicolumn{1}{c}{Testing Data} & Objects & MAP$\uparrow$           & FR$\uparrow$             \\ \midrule
		\multirow{2}{*}{$\mathcal{I}_5$+$\mathcal{O}_4$}     & $\mathcal{I}_{1-4}$                & 10          & 90.4          & \textbf{12.5}       \\
		& $\mathcal{O}_{1-3,5}$               & 28          & 49.7          & 16.0     \\ \midrule
		\multirow{2}{*}{$\mathcal{I}_5\left(50\right)$+$\mathcal{O}_{1-5}\left(50\right)$} & $\mathcal{I}_{1-4}$                
		& 10          & 90.4          & 12.2       \\
		& $\mathcal{O}_{1-5}\left(-50\right)$          & 33          & 81.9          & \textbf{17.1}     \\ \midrule
		\multirow{2}{*}{$\mathcal{I}_5$+$\mathcal{O}_{1-5}\left(100\right)$}    & $\mathcal{I}_{1-4}$                
		& 10          & 90.4          & 12.3       \\
		& $\mathcal{O}_{1-5}\left(-100\right)$         & 33          & 82.4          & 16.6     \\ \midrule
		\multirow{2}{*}{$\mathcal{I}_5$+$\mathcal{O}_{1-5}\left(200\right)$}    & $\mathcal{I}_{1-4}$                
		& 10          & 90.5          & 12.3       \\
		& $\mathcal{O}_{1-5}\left(-200\right)$         & 33         & 85.1          & 16.6      \\ \midrule
		\multirow{2}{*}{$\mathcal{I}_5$+$\mathcal{O}_{1-5}\left(300\right)$}    & $\mathcal{I}_{1-4}$                
		& \textbf{10}          & \textbf{90.5}          & 11.8       \\
		& $\mathcal{O}_{1-5}\left(-300\right)$         & \textbf{33}          & \textbf{85.8}          & 16.2     \\ \bottomrule
	\end{tabular}
	\caption{\label{tab:det_results} Detection results of different training sets.}
\end{table}

\begin{figure}[!htbp]
	\centering
	\includegraphics[width=0.6\linewidth]{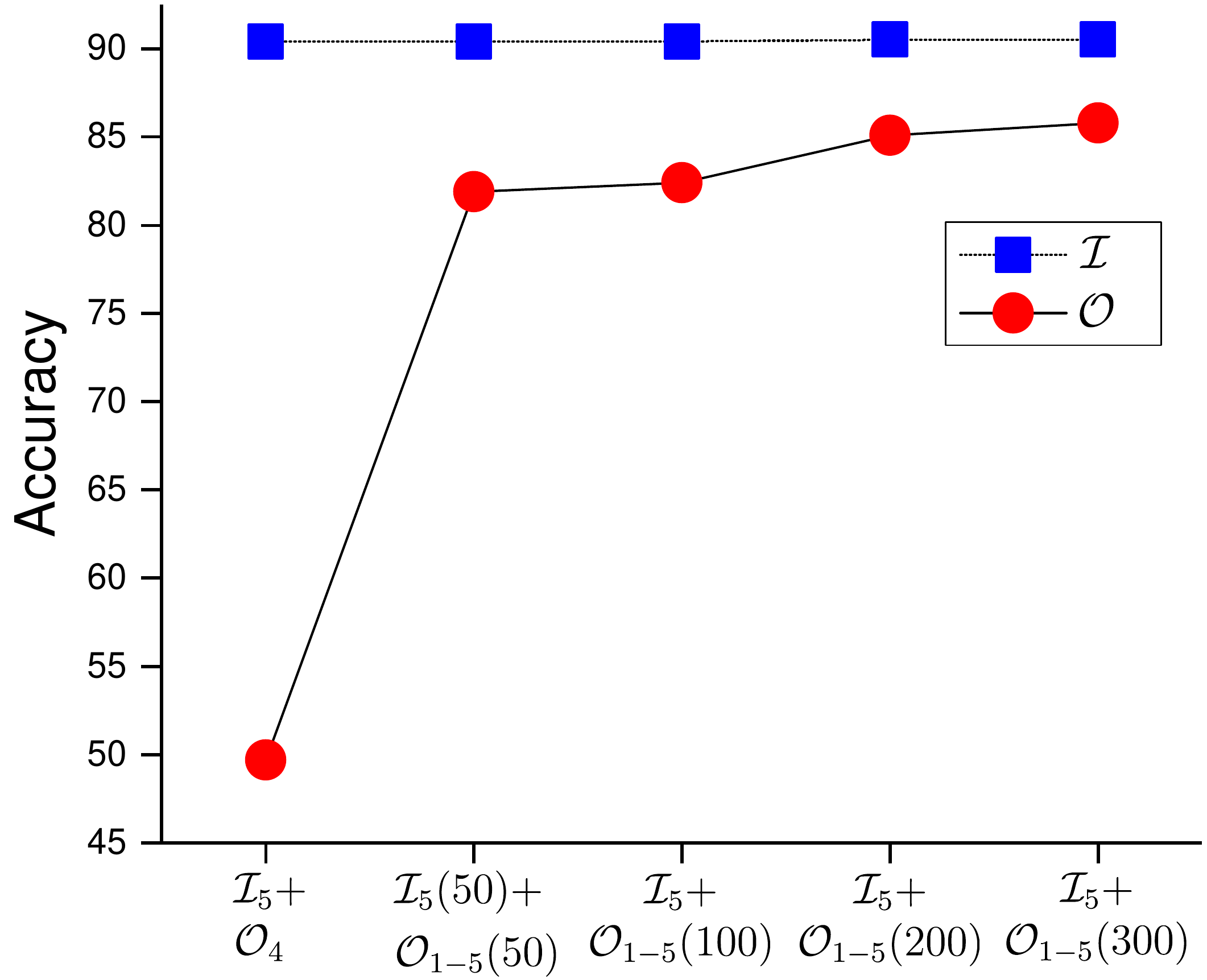}
	\caption{\label{fig:different_sets} Detection accuracy of different training sets.}
\end{figure}

In the subsequent experiments, we integrate the images of all outdoor scenes into the outdoor training set and dramatically improve the accuracy of outdoor testing.
Figure~\ref{fig:different_sets} clearly shows the effects of using different training sets. 
By further increasing in the number of images in outdoor videos, the detection accuracy of outdoor scenes improves slightly. 
For indoor environments, the detection accuracy is impervious to different training sets. 
Moreover, reducing the number of images to 50 ($\mathcal{I}_5$ has a total of 351 frames) does not reduce the detection accuracy. 
This shows that we need only a small number of training samples to achieve satisfactory results when the training and testing scenarios are the same.

The frame rate is around 12 FR for indoor videos and 16 FR for outdoor ones.
The factor of different image resolution should be accountable for this performance gap.
In practical applications, if accuracy is guaranteed, we tend to use smaller training sets to reduce labeling costs. 
Therefore, we use the model trained in “$\mathcal{I}_5\left(50\right)$+$\mathcal{O}_{1-5}\left(50\right)$” for comparison with the other methods in the comparative experiments.

\subsection{Results of multi-ant tracking}
\begin{figure}[!htbp]
	\centering
	\includegraphics[width=\linewidth, height=1\linewidth]{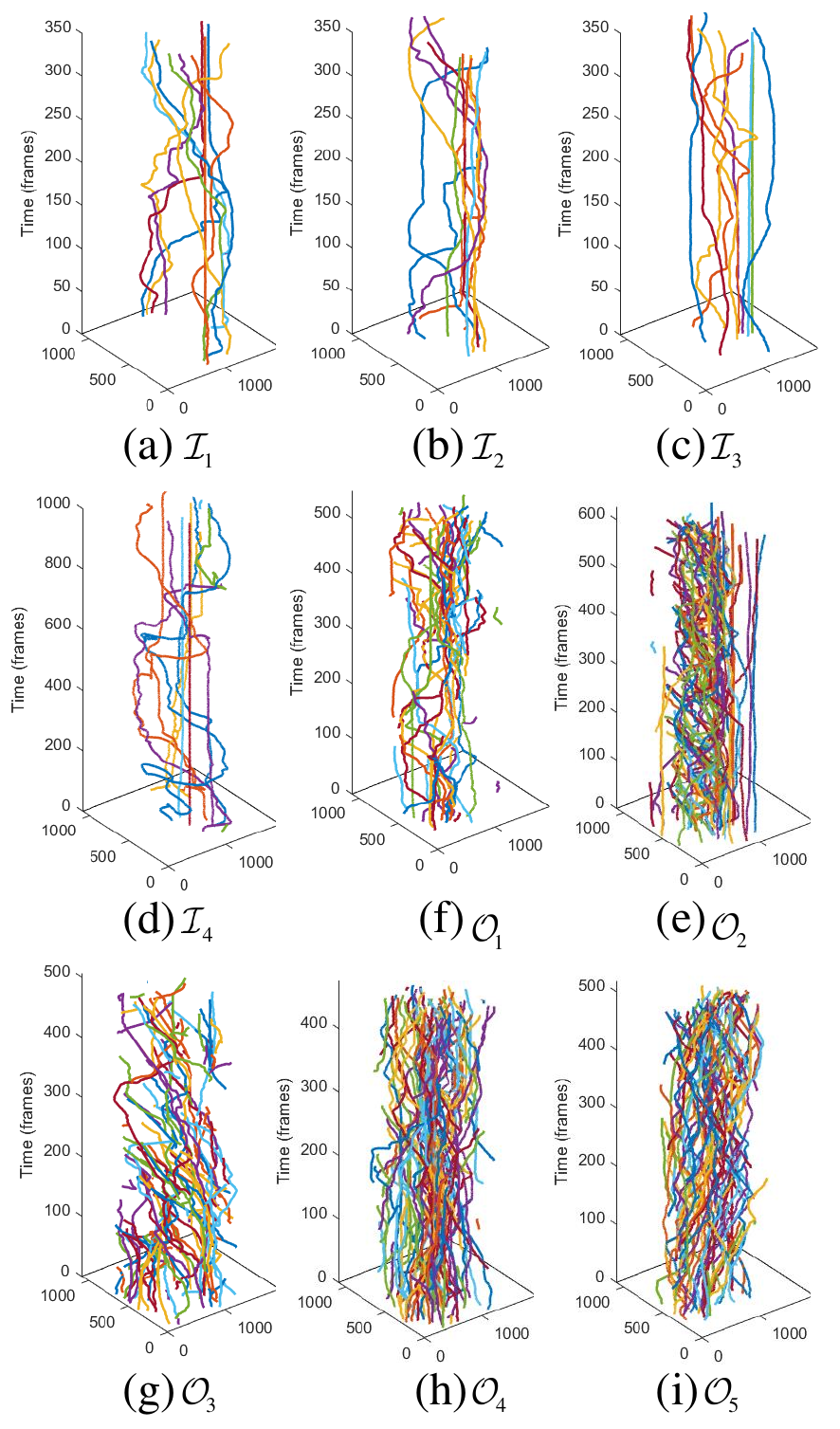}
	\caption{\label{fig:trajectories} Tracking trajectories in test videos. Horizontal axes indicate the pixel coordinates in an image. (a-d) indoor scenes. (e-i) outdoor scenes.}
\end{figure}

Based on the TBD paradigm, we use detection results as the input to the tracking framework. For offline training, we randomly select 50 labeled samples from $\mathcal{I}_5$ as the training set.
We visualized the tracking results in Figure~\ref{fig:trajectories}.

Table~\ref{tab:tracking_performance} shows the performance of online tracking.
After integrating the images of each outdoor video in the detection training set, our method gets 95\% mMOTA for indoor videos and over 80\% for outdoor videos. 
Additionally, mMOTP is around 80\% for both indoor and outdoor videos.
Notably, since the tracking performance depends on the detection result, the tracking task in $\mathcal{O}_{1-3,5}$ fails due to the low-quality detection (the second row in Table~\ref{tab:tracking_performance}).
Except for this failure case, the tracking performance is generally satisfactory considering that we only use 50 labeled samples from one indoor video.

\begin{table}[h]
	\centering
	{
	\begin{tabular}{@{}lcccccc@{}}
		\toprule
		\multicolumn{1}{c}{Testing Data} & FP$\downarrow$   & FN$\downarrow$   & IDS$\downarrow$   & mMOTA$\uparrow$ & mMOTP$\uparrow$ & FR$\uparrow$   \\ \midrule
		$\mathcal{I}_{1-4}$                & 236  & 621  & 21  & 89.9  & 79.9  & 36.5 \\
		$\mathcal{O}_{1-3,5}$               & \multicolumn{6}{c}{detection failure}                          \\ \midrule 
		$\mathcal{I}_{1-4}$                & 239  & 628  & 22  & \textbf{95.7}  & 81    & \textbf{38.9} \\
		$\mathcal{O}_{1-5}\left(-50\right)$          & 6078 & 7122 & 625 & 81.8  & 81.9  & 26.2 \\ \midrule 
		$\mathcal{I}_{1-4}$               & 260  & \textbf{617}  & \textbf{14}  & 95.7  & 81    & 38.7 \\
		$\mathcal{O}_{1-5}\left(-100\right)$         & 4867 & 6018 & 526  & 83.3  & 82.7  & \textbf{28.0} \\ \midrule 
		$\mathcal{I}_{1-4}$                & 228  & 709  & 17  & 95.4  & 81    & 36.3 \\
		$\mathcal{O}_{1-5}\left(-200\right)$         & 4289 & 3421 & 394  & 85.3  & 83    & 24.9 \\ \midrule  
		$\mathcal{I}_{1-4}$        & \textbf{224}  & 820  & 24  & 94.8  & \textbf{81.7}  & 35.4 \\
		$\mathcal{O}_{1-5}\left(-300\right)$         & \textbf{2644} & \textbf{3007} & \textbf{266}  & \textbf{85.8}  & \textbf{83.3}  & 26.4 \\ \midrule \hline
		$\mathcal{I}_{1-4}$/GT         & 22   & 23   & 8   & 99.6  & 92.4  & 35.2  \\
		$\mathcal{O}_{1-5}\left(-50\right)$/GT      & 1697 & 458  & 1064 & 96.2  & 92.4  & 25.9 \\ \bottomrule
	\end{tabular}}
	    \caption{\label{tab:tracking_performance} Tracking performance evaluation. The last two rows indicate that we use the ground truth of detection for tracking, which leads to a boost in tracking performance.}
\end{table}

The time cost of the tracking model is mainly incurred by generating 128-d feature vectors for each detection box. 
The average number of objects in outdoor videos is more than three times that in indoor videos. 
As for runtime time, FR reaches over 35 in indoor videos and more than 24 in outdoor videos.

We add a set of comparative experiments in the last two rows of Table~\ref{tab:tracking_performance}. 
We directly use manually-labeled detection boxes for tracking and compare the detection results on the $\mathcal{I}_{1-4}$ and $\mathcal{O}_{1-3,5}$.
Both mMOTA and mMOTP have been dramatically improved.
This implies that an increase in detection accuracy could further boost the tracking performance of our framework.

\subsection{Comparative experiments}
\label{sec:comparative}

\begin{figure}[!htbp]
	\centering
	\includegraphics[width=\linewidth]{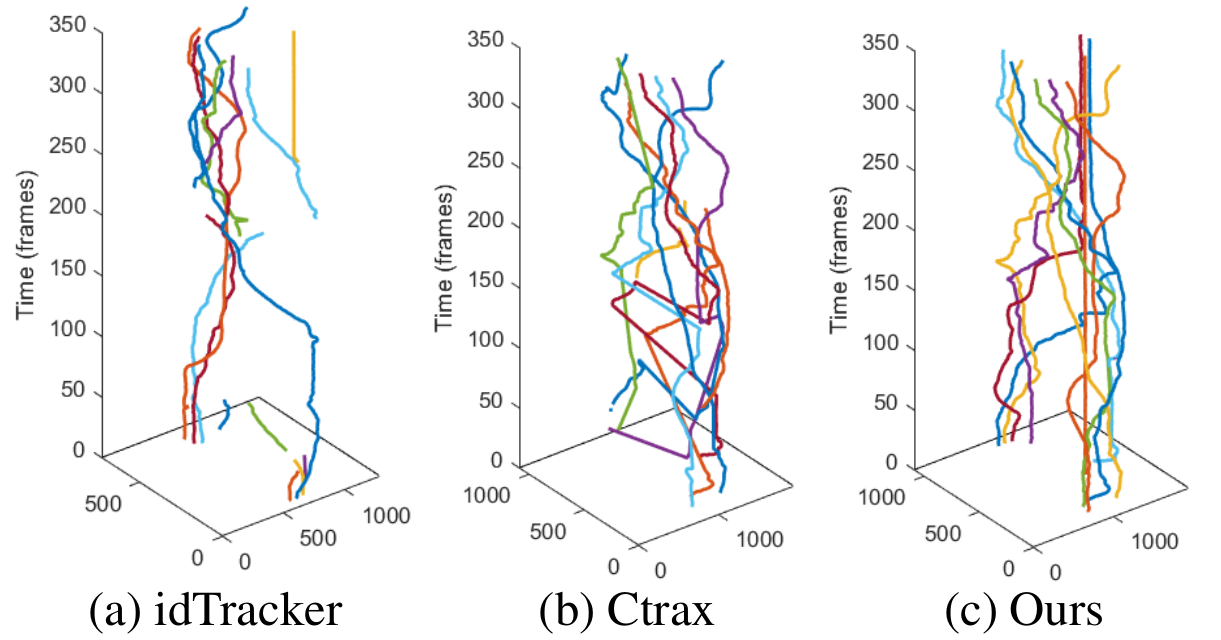}
	\caption{\label{fig:spatial_temporal} Comparison of tracking performance in spatial-temporal dimension ($\mathcal{I}_1$). Horizontal axes indicate the pixel coordinates in an image.}
\end{figure}

There are two widely used insect tracking software: idTracker \cite{perez2014idTracker} and Ctrax \cite{chiron2013detecting}. 
idTracker needs to specify the number of objects before tracking, to create a reference image set for each object. 
Meanwhile, Ctrax assumes that objects will rarely enter and leave the arena. 
Thus, they are both not capable of tracking in outdoor scenes because of the variable number of ants. 
Therefore, we compare these two methods only in videos depicting indoor scenes.
idTracker needs to specify the number of objects before tracking, in order to create a reference image set for each object. 
To compare them with our method, we convert their representations into square boxes as our ground truth.
Table~\ref{tab:comparison} and Figure~\ref{fig:spatial_temporal} shows the tracking results.
In addition to a significant improvement of tracking accuracy, our method is 6 and 10 times faster than idTracker and Ctrax (see the column of FR).

\begin{table}[h]
	\centering
	{
	\begin{tabular}{cccccccc}
		\toprule
		Method    & FP$\downarrow$   & FN$\downarrow$   & IDs$\downarrow$ & FM$\downarrow$  & mMOTA$\uparrow$ & mMOTP$\uparrow$ & FR$\uparrow$  \\ \hline
		idTracker & 881  & 8479 & 83  & 432 & 54   & 77.4 & 1.3 \\
		Ctrax     & 2832 & 5646 & 110 & 349 & 58.2 & 79.7 & 0.8 \\
		Ours      & \textbf{239}  & \textbf{628}  & \textbf{22}  & \textbf{189} & \textbf{95.7} & \textbf{81.1} & \textbf{8.7} \\ \bottomrule
	\end{tabular}}
	\caption{\label{tab:comparison} Comparison of tracking results on videos $\mathcal{I}_{1-4}$.}
\end{table}


\begin{figure}[!htbp]
	\centering
	\includegraphics[width=\linewidth]{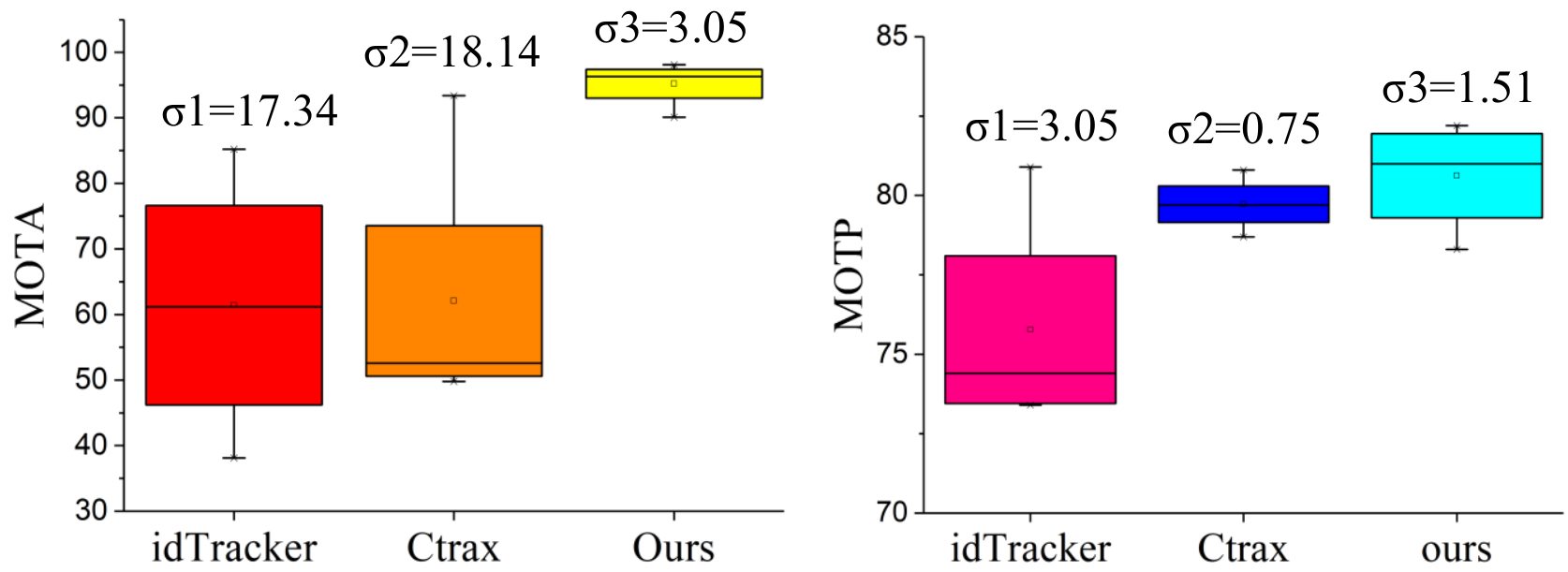}
	\caption{\label{fig:comparison} Comparison of tracking results for indoor scenes.}
\end{figure}

We further compare tracking accuracy of idTracker and Ctrax across different indoor videos, as Figure~\ref{fig:comparison} shows.
The large variance of idTracker's performance is affected by the number of static ants, which will cause missing tracking.
Ctrax proves to be robust but with a lower accuracy compared with our method.

\section{Discussion}

\subsection{Comparison of methods}

In the previous section, we compared two methods of insect tracking, idTracker and Ctrax.
idTracker uses the intensity and contrast of the foreground segmented area to extract appearance features and construct a reference image set for each individual. 
However, it can not track motionless individuals.  
Figure~\ref{fig:spatial_temporal}(a) shows that only a minority group of ants are successfully tracked over the period of video.
Further, there are some trajectory fragments due to the limitations of the foreground segmentation model for multiple objects.

Compared to our results, the trajectories of Ctrax are incomplete.
This indicates that there are more FN, as Figure~\ref{fig:spatial_temporal}(b) shows. 
Ctrax requires a sharp contrast between object and background. 
The ants passing through the overexposed areas in the scene will be ignored. 
Additionally, Ctrax assumes that the motion of the object obeys the linear distribution. 
However, the ants’ movement is nonlinear, and their speed and direction might change abruptly, causing IDS in Ctrax.

Our method classifies and regresses twice to locate ants accurately. 
During the tracking stage, we use the historical appearance sequence as a reference and update it frame by frame. 
Compared with idTracker, our method effectively solves the long-term and short-term dependence of motion states, thereby reducing FM.
Despite that we also assume the linear distribution of motion states, they are used only to filter impossible associations, and have nothing to do with association cost.
We take the appearance distance between trajectories and detection boxes as association cost, thus the model is robust even when the ant movement is complicated.
We take the appearance measure between trajectories and detection boxes as the association cost, thus the model is robust even when the ant movement is complicated.

\subsection{Failure cases}
\label{sec:failure}

\subsubsection{Limitations of detection framework}
\begin{figure}[!htbp]
	\centering
	\includegraphics[width=\linewidth]{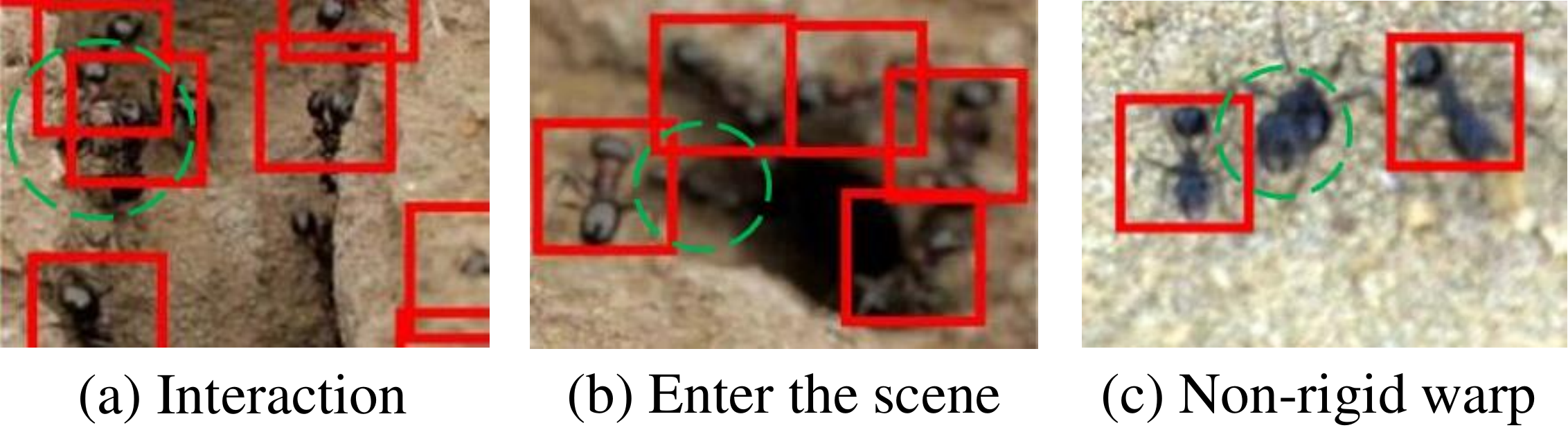}
	\caption{\label{fig:failed_detection} Examples of failed detection in outdoor scenes.}
\end{figure}

The number of ants in outdoor scenes is on average 33 per frame. 
It is also typical for ants to involve close body contact with each other for the purpose of information sharing.
Naturally, their extremely-close interactions are highly likely to cause mis-detection (Figure~\ref{fig:failed_detection}(a)).
Additionally, entrances and exits of ants in outdoor scenes are more prone to mis-detection (Figure~\ref{fig:failed_detection}(b)). 
Moreover, the dramatically non-rigid deformation of ants is also a factor causing the detection failure (Figure~\ref{fig:failed_detection}(c)).
These three scenarios are all challenging cases that deserve our future efforts.

\subsubsection{Limitations of tracking framework}
\begin{figure}[!htbp]
	\centering
	\includegraphics[width=\linewidth]{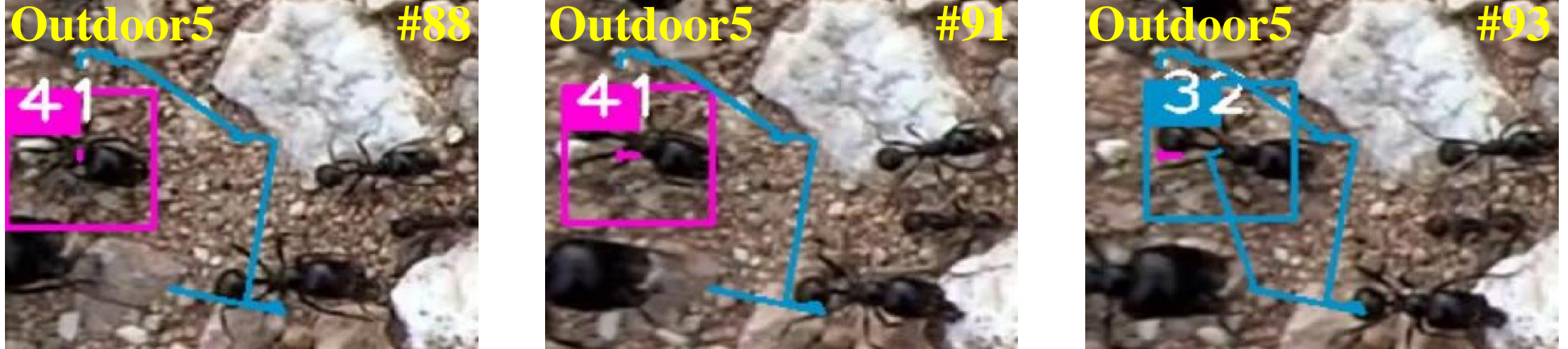}
	\caption{\label{fig:drift} Drift at the scene boundary. A newly-entered Ant No.41 is mis-associated with an existing Trajectory No.32.}
\end{figure}

According to Figure~\ref{fig:drift}, Ant No.41 entered the scene at Frame No.88. 
Coincidentally Ant No.32 left the scene at an adjacent region, but its trajectory was not deleted. 
At Frame No.93, Ant No.41 drifted to Trajectory No.32.
This defect is caused by insufficient appearance descriptors stored in the gallery of Ant No.41, and it moved near the exit location of another ant.
This kind of mis-association occurs at the image boundary and accounts for the majority of IDS and FM in our experiments. 
However, when ants move inside the scope of both indoor and outdoor scenes, our method can accurately track multiple ants simultaneously for a long time, as Figure~\ref{fig:tracking_results} shows.

\section{Conclusion}
We proposed a complete detection and tracking framework based on deep learning for ant colony tracking. 
In the detection stage, we adopted a two-stage object detection framework for the detection task. 
We also use a ResNet model to obtain ant appearance descriptors for online associations. 
Next, we combined appearance and motion information for the tracking task. 
The experimental results demonstrated that our method outperformed two mainstream insect tracking models in terms of accuracy, precision, and speed. 
Particularly, our work shows its advantage in robustly detecting and tracking ant colonies in outdoor scenes, which is rarely reported in existing literature.
We believe our method could serve as an effective tool for high-throughput swarm behavior analysis of ant colonies, leading to the development of embodied intelligence.

In future research, we aim to achieve more robust detection.
For example, by exploring additional information of ants' skeletal structure, we can potentially solve the aforementioned failure case of close interaction and nonrigid deformation problem. 
We also plan to improve the generalization ability of our detection and tracking frameworks so that it is applicable to a wide range of outdoor environments.

\section*{Conflict of Interest Statement}

The authors declare that the research was conducted in the absence of any commercial or financial relationships that could be construed as a potential conflict of interest.

\section*{Author Contributions}

All authors listed have made substantial, direct, and intellectual contribution to the work; they have also approved it for publication.
In particular, MW, SG and XC contributed to the design of this work; MW, SG and XC contributed to the writing of the manuscript; XC designed and implemented the multi-ant tracking framework; XC conducted the experiments, and analyzed the results.


\section*{Acknowledgments}
This work was supported by the Natural Science Foundation of Fujian Province of China (No. 2019J01002).

\section*{Data Availability Statement}
The training and testing datasets used for this study can be found in the \href{https://data.mendeley.com/datasets/9ws98g4npw/3}{ANTS--ant detection and tracking}.

\bibliographystyle{Frontiers-Harvard} 
\bibliography{ref}

\end{document}